# Evaluating Multiple Instance Learning Strategies for Automated Sebocyte Droplet Counting

Maryam Adelipour[1], Gustavo Carneiro[2], Derrick Adams,[3] Kee K. Kim,[4] Jeongkwon Kim[1*]

[1]Department of Chemistry, Chungnam National University, Daejeon, 34134, Republic of Korea

[2]Centre for Vision, Speech and Signal Processing, University of Surrey, UK

[3]Department of Chemical and Biomolecular Engineering,Korea Advanced Institute of Science and Technology (KAIST), 291 Daehak-ro, Yuseong-gu, Daejeon 34141, Republic of Korea

[4]Department of Biochemistry, Chungnam National University, Daejeon, Republic of Korea

*Correspondence: Jeongkwon Kim jkkim48105@cnu.ac.kr

## Abstract

Sebocytes are lipid-secreting cells whose differentiation is marked by the accumulation of intracellular lipid droplets, making their quantification a key readout in sebocyte biology. Manual counting is labor-intensive and subjective, motivating automated solutions. Here, we introduce a simple attention-based multiple instance learning (MIL) framework for sebocyte image analysis. Nile Red-stained sebocyte images were annotated into 14 classes according to droplet counts, expanded via data augmentation to about 50,000 cells. Two models were benchmarked: a baseline multi-layer perceptron (MLP) trained on aggregated patch-level counts, and an attention-based MIL model leveraging precomputed ResNet-50 feature embeddings with trainable instance weighting. Experiments using five-fold cross-validation showed that the baseline MLP achieved more stable performance (mean MAE = 5.6) compared with the attention-based MIL, which was less consistent (mean MAE = 10.7) but occasionally superior in specific folds. These findings indicate that simple bag-level aggregation provides a robust baseline for slide-level droplet counting, while attention-based MIL requires task-aligned pooling and regularization to fully realize its potential in sebocyte image analysis.

## Introduction

Sebaceous glands are lipid-secreting skin appendages typically associated with hair follicles, together forming the pilosebaceous unit. Their activity depends on sebocytes, which transition from proliferative basal cells to differentiated cells that accumulate cytoplasmic lipid droplets and ultimately disintegrate through holocrine secretion, releasing sebum. This complex mixture of triglycerides, wax esters, squalene, and cholesterol reinforces the skin barrier providing antimicrobial and antioxidative protection. The quantity and composition of sebum vary with species and age, and disturbances in sebocyte function are closely associated with dermatological disorders. Excessive secretion and altered lipid fractions are central to acne vulgaris, while sebaceous gland atrophy characterizes several forms of scarring alopecia (1, 2). The human-specific presence of large sebaceous follicles, absent in most other mammals, underscores the importance of studying sebocyte biology directly in human systems. To facilitate such investigations, immortalized sebocyte cell lines have been established as reproducible in vitro models for dissecting sebaceous lipogenesis and evaluating therapeutic strategies (2). A hallmark of sebocyte maturation into fully functional, sebum-secreting cells is the accumulation of intracellular lipid droplets, which represent reliable morphological and functional markers of differentiation (3). Quantitative analysis of these droplets in microscopy images is therefore essential for sebocyte research; yet manual assessment is labor-intensive, subjective, and inconsistent, highlighting the need for automated and robust image analysis approaches. Recent developments in artificial intelligence provide promising solutions to address this need.

Convolutional neural networks have shown outstanding performance in medical image segmentation, classification, and feature extraction, enabling automated analysis of large-scale image datasets with accuracy and reproducibility that often surpass manual expert evaluation (4-6). These advances highlight the potential of AI to automate microscopy-based diagnostics and experimental workflows, which are traditionally time-consuming and prone to observer variability. Applications span from fundamental biological research, including the study of cell differentiation and intracellular organization, to clinical pathology, where AI-driven image analysis supports disease detection and prognostic assessment using histological slides (7). Moreover, the increasing availability of open-source frameworks and pretrained models has made deep learning more accessible to researchers in cell and developmental biology, thereby lowering the technical barriers for non-specialists and fostering broader adoption (8). This growing ecosystem of tools has paved the way for scalable and reproducible bioimage analysis, yet specific applications,—such as the automated quantification of lipid droplets in sebocytes as a proxy for cellular differentiation,—remain underexplored, representing an important opportunity for methodological innovation.

In this work, sebocyte droplet quantification is formulated as a bag-level, multi-target regression problem: for each microscopy slide, a set of patch features is provided as input, and a 14-dimensional count vector is produced as output, indicating the number of cells in each droplet-count class. Performance is summarized using mean absolute error (MAE) under a five-fold cross-validation protocol with three random seeds. Two complementary strategies are evaluated on identical features and splits. A bag-level multilayer perceptron (MLP) aggregates slide-level information to directly regress the 14-dimensional count vector, providing a robust baseline. An attention-based multiple-instance learning (MIL) model is then used to learn instance weights that emphasize informative regions while producing the same slide-level output. This comparison is designed to test whether attention mechanisms provide advantages over standard bag-level aggregation for droplet counting, without reliance on pixel-level annotations.

This study aims to benchmark sebocyte droplet counting as bag-level multi-target regression using a five-fold, three-seed protocol, by comparing a bag-level MLP against an attention-based MIL on identical features and splits. The analysis prioritizes accuracy and stability, demonstrating stable MAE for the MLP and fold-specific competitiveness, while exhibiting higher variance for the attention-based MIL, thereby clarifying the areas where attention and pooling have the most tremendous impact.

## Methodology

### Dataset Preparation

Sebocytes were initially expanded in their specific medium (Human Sebocyte Cell Culture Complete Growth Media with serum and antibiotics, Cellprogen). For differentiation, the cells were cultured in Sebomed basal medium (Sigma, USA) supplemented with 1% FBS and 1% penicillin–streptomycin under standard conditions (37 °C, 5% $CO_2$) for 7 days. Cells were seeded at a density of 30,000 per well into 6-well culture plates (SPL, South Korea). Following differentiation, intracellular lipid droplets were visualized using Nile Red staining according to the manufacturer's instructions (Abcam, ab228553). Briefly, cells were incubated with Nile Red for 30 min at 37 °C, washed with PBS, and imaged on a fluorescence microscope (Olympus IX73, Japan) at 20× magnification. Images were saved as JPEG (Baseline), sRGB, 8-bit/channel at a native resolution of 4096 × 2160 p ixels using a DIXI USB3X9 camera. Across sessions, EXIF metadata indicated exposure times of ~0.34–0.44 s. The green channel was used for droplet counting due to superior signal-to-background contrast. In total, 80 images were acquired over a 7-day culture period and compiled into a dataset spanning different stages of sebocyte differentiation. All images were annotated in Label Studio. Each sebocyte was delineated with a polygonal mask and assigned to one of 14 ordinal classes indexed 0 through 13 according to the number of visible lipid droplets. By definition, class 0 denotes cells without droplets; class 1 denotes cells with one droplet; continuing in unit increments up to class 12, which denotes twelve droplets; and class 13 denotes thirteen or more droplets (i.e., >12). This scheme provides a consistent, count-based target for supervised learning at the slide (bag) level. For droplet counting, a "standard droplet" was defined to reduce bias from size heterogeneity. Selected images were converted to 8-bit grayscale (ImageJ/Fiji: Image > Type > 8-bit), and a global intensity threshold was adjusted to isolate lipid droplets (Image > Adjust > Threshold). The resulting binary mask was used to identify connected components, from which representative droplets were selected for measurement. A droplet with an area near the median of typical droplets was chosen as the standard, cropped, and saved as a visual ruler for use during manual counting. During annotation, tiny speckles visibly below this standard were treated as noise and ignored; droplets comparable in size to the standard were counted as one; and larger droplets were assigned an equivalent count proportional to their area (approximately the droplet area divided by the standard-droplet area), using the visual ruler as guidance. Threshold settings and example snapshots of the standard droplet and equivalent-count assignments were archived to support reproducibility. After annotation of the 80 original images, data augmentation was applied to increase dataset diversity. Augmentation included brightness adjustment (0.8× and 1.2× relative to the original images) and Gaussian blurring (kernel size = 3), generating three additional datasets. These augmentations were chosen to reflect realistic variations encountered in microscopy, such as differences in illumination or minor lens adjustments, which can affect image brightness and sharpness. To ensure traceability, all augmented images were saved with unique suffixes appended to the original filenames, and corresponding JSON annotation files were generated with the same suffixes. Individual annotation files were then merged into a single dataset-level JSON file by sequentially appending entries to the list structure of the first file. In total, 80 original images yielded 12,618 annotated sebocytes (≈158 cells per image) across 14 classes. The most populated class corresponded to cells containing more than 12 lipid droplets (3,911 cells), while the least represented class was cells with 11 droplets (213 cells). After applying three types of augmentation, the dataset expanded to 320 images comprising approximately 50,000 annotated sebocytes.

**Model architecture**

The attention-based multiple instance learning (MIL) model was implemented following the formulation of Ilse et al. (2018), and adapted from the Generalist Pathology Foundation Model (GPFM) framework (9). Each microscopy slide was represented as a bag of precomputed patch-level embeddings, where 512×512 image patches were encoded offline using a ResNet-50 backbone pretrained on ImageNet. These embeddings were kept fixed during training and served as the input to the MIL model. Within each bag, instance embeddings were aggregated through a trainable attention module, which assigned normalized weights ($\alpha_i$) to individual patches, allowing the model to focus on regions most informative for lipid droplet content. The resulting attention-weighted bag representation was passed through a fully connected regression head to produce 14 output channels ($count_0$–$count_{13}$), corresponding to the per-cell lipid droplet count distribution. A softplus activation was used to ensure non-negative outputs. Training was performed using a channel-weighted mean squared error loss on log1p-transformed predicted and target count vectors, where per-channel weights were computed as the inverse of the mean target frequency to mitigate imbalances across droplet count categories. Because the feature extractor was frozen and only the attention module and regression head were trained, this setup constitutes a weakly supervised slide-level learning formulation that does not require cell-level or pixel-level annotations, while preserving interpretability through attention maps that highlight patch regions contributing to lipid accumulation patterns (Figure 1A).

## Baseline model

To establish a reference framework, a baseline multi-layer perceptron (MLP) regression model was implemented. Similar to the MIL setup, precomputed patch-level embeddings extracted from the fixed ResNet-50 backbone were used. However, instead of learning instance-level importance through attention, all patch embeddings belonging to the same slide were mean-pooled into a single slide-level feature vector prior to training. Thus, feature aggregation was performed externally, without trainable weighting. The resulting slide-level representation was passed through a shallow MLP composed of two fully connected layers: an input layer followed by a hidden layer with 256 units, ReLU activation, and dropout ($p = 0.2$), and an output layer producing 14 regression outputs ($count_0$–$count_{13}$), each corresponding to a lipid droplet count category. A softplus activation was applied to ensure non-negative outputs. The baseline model was trained using the same log1p-transformed, channel-weighted MSE loss used for the MIL model, enabling a fair and directly comparable evaluation (Figure 1B). This baseline provided a simplified yet informative benchmark, allowing the impact of trainable attention-based instance aggregation in the MIL framework to be quantitatively isolated and assessed.

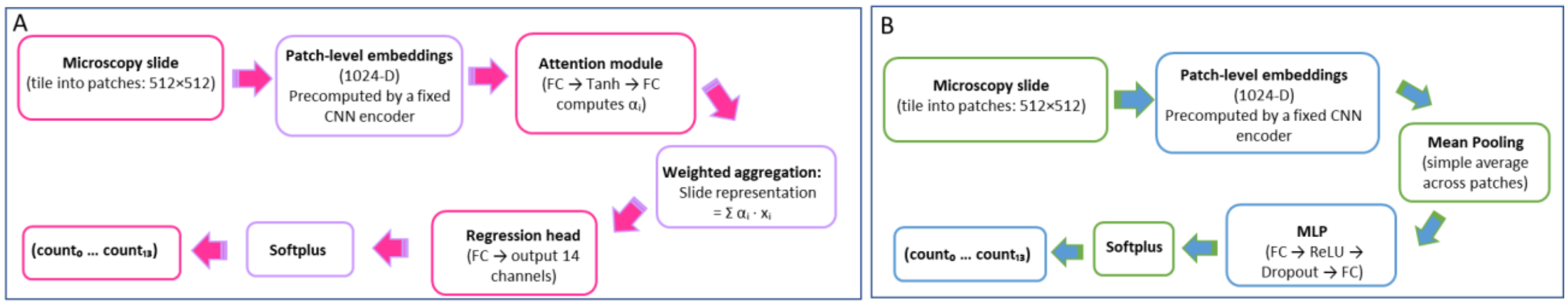

**Figure 1. Overview of the two modeling strategies used for slide-level lipid droplet quantification.** A. Attention-based MIL model: Each microscopy slide is tiled into 512×512 patches, and patch-level embeddings (1024-D) precomputed by a fixed CNN encoder are used as input instances. A trainable attention module assigns normalized weights ($\alpha_i$) to patches, enabling weighted aggregation into a single slide-level representation, followed by a fully connected regression head with Softplus activation to generate 14 non-negative droplet-count outputs. B. Baseline MLP model: The same precomputed patch embeddings are mean-pooled across all patches to form a slide-level feature vector. This representation is passed through a shallow MLP (FC → ReLU → Dropout → FC), followed by Softplus to predict the same 14

output channels. This model serves as a non-attention benchmark for evaluating the contribution of learned instance weighting.

### Training setup

All experiments were conducted under a five-fold cross-validation (CV) scheme to ensure robust and unbiased performance estimation. For each fold, distinct slide-level metadata tables (train/validation/test CSVs) were generated such that all patches originating from the same microscopy slide—along with every augmented variant derived from that slide—were confined to a single split. This strict grouping strategy eliminated data leakage between sets, ensuring that neither original nor augmented samples from the same source slide appeared across training, validation, or test subsets. Data augmentation, including random rotations, horizontal/vertical flips, and intensity jitter, was applied after fold assignment, thereby preserving independence between folds while enhancing model generalization. Training was performed on a single NVIDIA V100 GPU with a bag-level batch size of 1, while larger effective batches were simulated via gradient accumulation (×4). Gradient clipping ($\|g\|_2 \leq 1.0$) was used to stabilize optimization. Both the attention-based MIL model and the baseline MLP model were trained using the Adam optimizer (learning rate = $1 \times 10^{-4}$, weight decay = $3 \times 10^{-4}$) for up to 200 epochs, with early stopping (patience = 20 epochs) based on the validation mean absolute error (MAE). To improve numerical stability, droplet count vectors were $\log_{1+}$-transformed prior to loss computation and inverse-transformed for evaluation. The final model performance was summarized as the mean ± standard deviation across all folds and random seeds, ensuring a fair and reproducible comparison between the MIL and MLP architectures trained under identical experimental conditions.

### Evaluation metrics

Model performance was quantitatively assessed at the slide level using mean squared error (MSE) and MAE between the predicted and ground-truth droplet count distributions. To ensure numerical stability, all predictions were generated in the $\log_{1+}$-transformed space and subsequently mapped back to the original scale before computing evaluation metrics. During training, validation MAE served as the primary criterion for early stopping and model checkpoint selection. Final test performance was reported as the mean ± standard deviation of both MAE and MSE across the five cross-validation folds and random seeds, ensuring statistically robust comparison between the MIL and MLP models.

### Implementation details

All experiments were implemented in PyTorch (v2.2) and executed on a single NVIDIA V100 GPU within a controlled HPC environment. Mixed-precision training with automatic loss scaling (AMP) was enabled to improve computational efficiency and memory utilization. The feature extraction backbone (ResNet-50) was frozen during training, ensuring that only the attention and regression/classification layers were updated. Both the MIL and baseline MLP models were trained under identical optimization and logging configurations to ensure reproducibility. Gradient accumulation, deterministic random seeds, and consistent data loaders were employed to maintain numerical reproducibility across different folds and random initializations. All code and configurations were version-controlled and executed within a shared Conda environment that contained standardized dependencies for PyTorch, NumPy, and scikit-learn.

## Results

### Annotation and dataset preparation

Manual annotation of sebocyte images was performed using Label Studio, where each cell was segmented and assigned to one of 14 predefined droplet-count classes (count_0–count_13) based on the number of intracellular lipid droplets (from 0 to ≥12). The annotated masks provided instance-level localization and quantitative labels that were later aggregated into slide-level count distributions for model training and evaluation. Representative examples of the annotated images are shown in Figure 2, where each color-coded mask corresponds to a distinct droplet-count class. These annotations served as the ground truth for training.

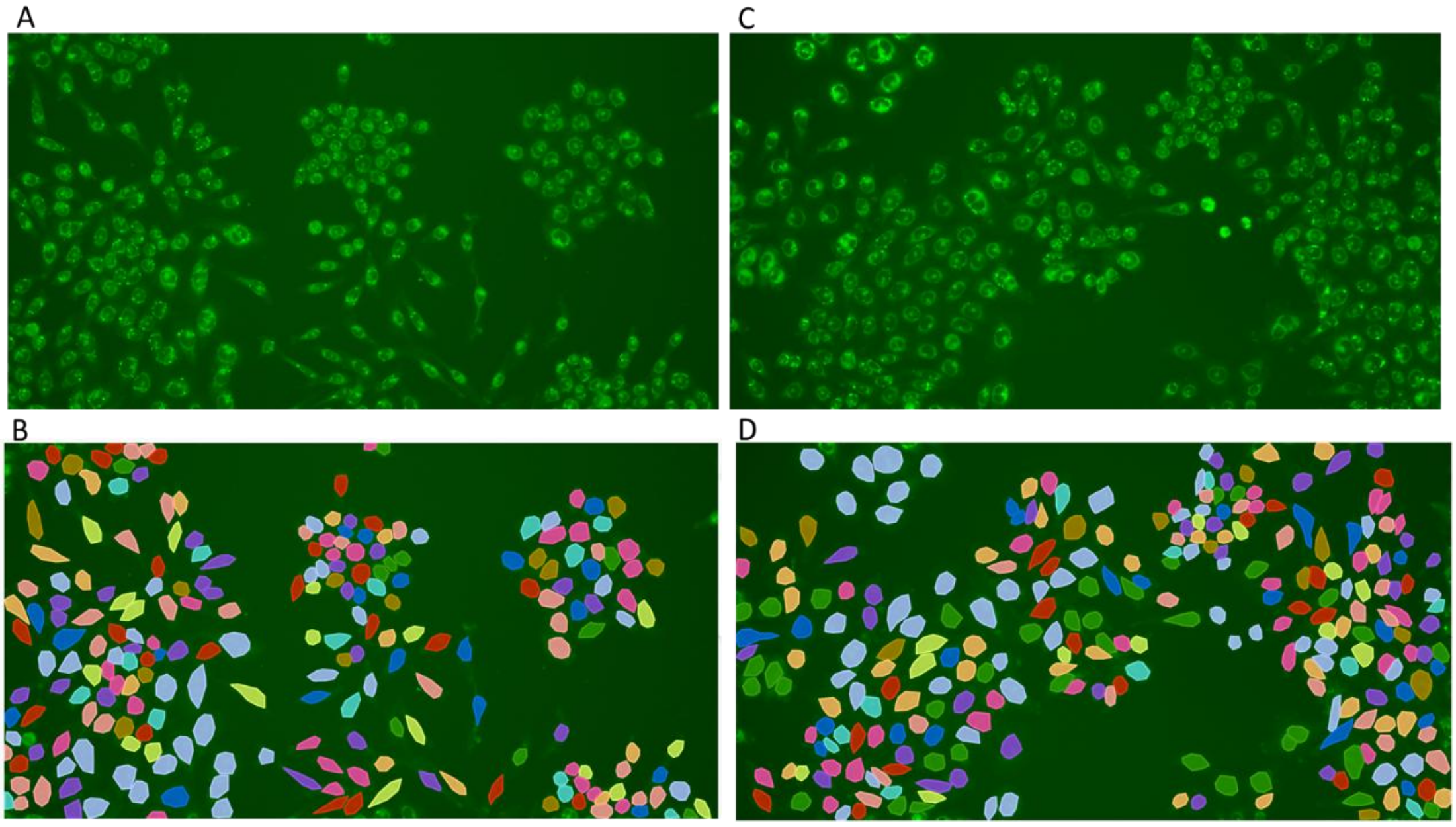

**Figure 2. Annotation of sebocyte microscopy images stained with Nile Red.** (A&C) Original fluorescence images of sebocytes showing intracellular lipid droplets with varying abundance. (B&D) Manual annotations created in Label Studio, where each cell is segmented and assigned a color corresponding to its droplet-count class (count_0–count_13).

### Data augmentation

To increase the size and variability of the dataset, image augmentation was performed. The augmentation parameters were carefully selected so that the overall quality and biological features of the images were not significantly altered. This ensured that key morphological structures, such as lipid droplets within sebocytes, remained clearly visible and biologically interpretable (Figure 3).

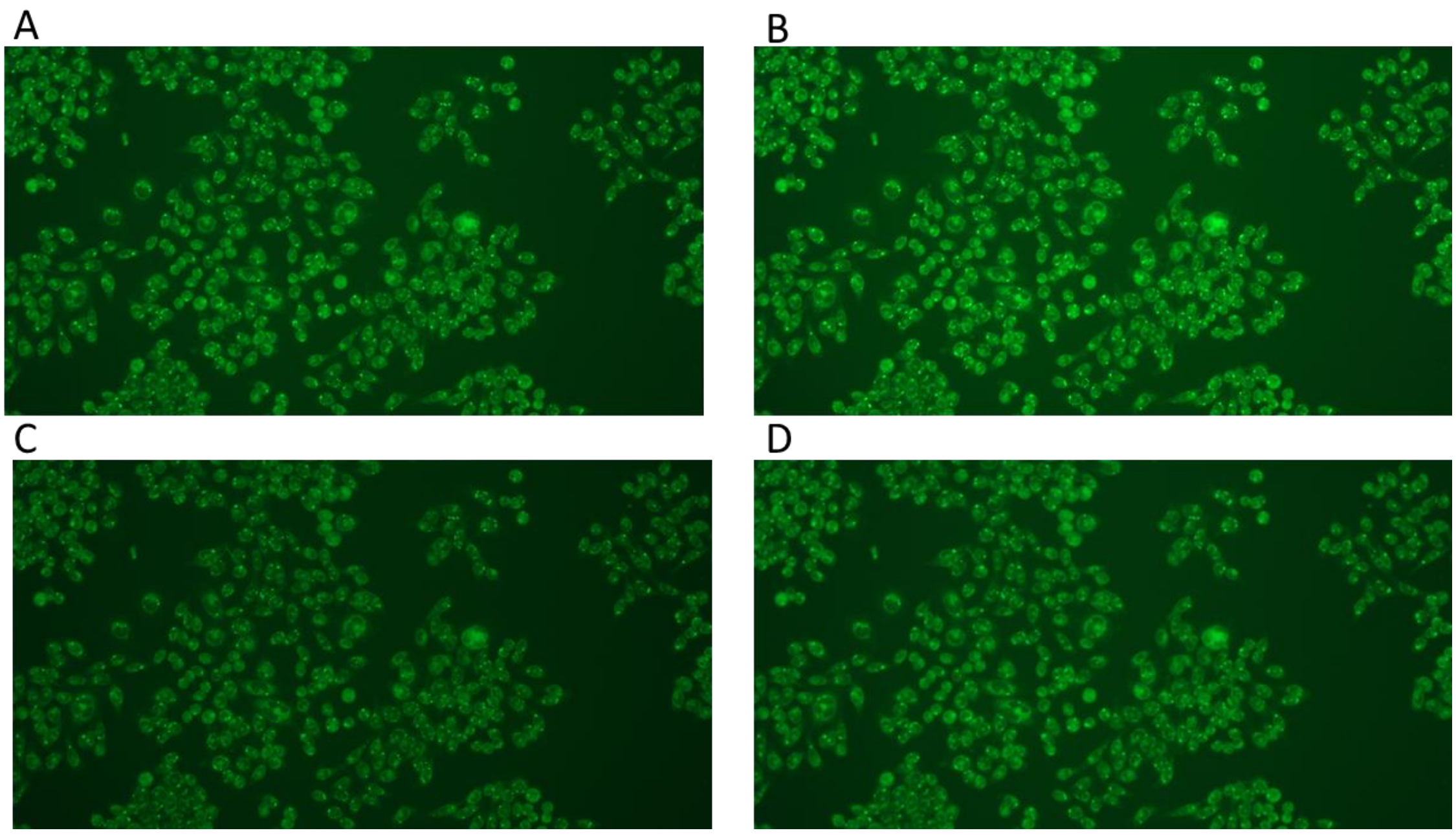

**Figure 3. Data augmentation applied to sebocytes stained with Nile Red**. A) Original image showing sebocytes with intracellular oil droplets. B) Image A with increased brightness (factor 1.2). C) Image A with reduced brightness (factor 0.8). D) Image A with applied blurriness (kernel size 3).

**Baseline MLP performance**

To establish a reference framework, a baseline MLP model was trained on aggregated patch-level count features. Unlike the attention-based MIL architecture, which adaptively weights individual instances, this baseline operates directly on bag-level representations, providing a more straightforward but consistent comparison. The model was trained for 200 epochs using three random seeds (11, 22, and 33) and evaluated across five cross-validation folds. As summarized in Table 1, the baseline MLP achieved a mean test MAE of 6.476 ± 1.811 across all seeds and folds, with per-seed averages of 6.456 ± 1.799 (seed 11), 6.483 ± 1.822 (seed 22), and 6.489 ± 1.812 (seed 33). The corresponding validation MAE values were 11.064 ± 3.074, 11.248 ± 3.147, and 11.087 ± 3.073, respectively, yielding an overall validation MAE of 11.133 ± 3.099. These results demonstrate the robustness and reproducibility of the baseline MLP across multiple random initializations, providing a stable non-attentive benchmark under identical optimization settings. Subsequent analyses isolate the added contribution of instance-level attention mechanisms in the MIL-based architecture.

**Table 1.** Validation and test performance (MAE) of the baseline MLP model across folds and seeds.

| Seed | Fold | ValMAE | TestMAE | Mean ValMAE (±SD) | Mean TestMAE (±SD) |
|---|---|---|---|---|---|
| 11 | 0 | 10.9999 | 4.2882 | | |
| 11 | 1 | 13.8373 | 7.8772 | | |
| 11 | 2 | 5.1885 | 9.1906 | | |
| 11 | 3 | 12.5734 | 5.5176 | | |
| 11 | 4 | 12.7194 | 5.4062 | 11.063700 ± 3.073612 | 6.455960 ± 1.798882 |
| 22 | 0 | 11.3176 | 4.2845 | | |
| 22 | 1 | 13.9088 | 7.9481 | | |
| 22 | 2 | 5.1765 | 9.2326 | | |
| 22 | 3 | 12.9853 | 5.5764 | | |

| 22 | 4 | 12.8499 | 5.3723 | 11.247620 ± 3.147447 | 6.482780 ± 1.821856 |
|---|---|---|---|---|---|
| 33 | 0 | 11.0074 | 4.3587 | | |
| 33 | 1 | 13.7623 | 7.8203 | | |
| 33 | 2 | 5.2021 | 9.3212 | | |
| 33 | 3 | 12.6959 | 5.5705 | | |
| 33 | 4 | 12.7662 | 5.3763 | 11.086780 ± 3.072585 | 6.489400 ± 1.811830 |
| Overall | --- | --- | --- | 11.132700 ± 3.099159 | 6.476047 ± 1.810938 |

**Attention-based MIL model performance**

To assess the contribution of instance-level attention, the adapted GPFM-based MIL model was trained using the same aggregated patch-level features under three random seeds (1, 2, and 3) and evaluated across five cross-validation folds. As summarized in Table 2, the model achieved a mean validation MAE of 20.31 ± 4.96 and a mean test MAE of 9.16 ± 3.14 across all seeds and folds. The per-seed averages were 20.04 ± 5.39 / 9.24 ± 3.22 (seed 1), 20.21 ± 5.12 / 9.24 ± 3.30 (seed 2), and 20.69 ± 5.52 / 9.00 ± 3.62 (seed 3) for validation and test MAE, respectively. Compared with the baseline MLP (overall 6.48 ± 1.81 MAE on test sets), the attention-based MIL model exhibited higher absolute error and greater variability across folds. This suggests that, under the present data regime, the attention mechanism did not yield consistent performance gains, likely due to the limited number of training instances per bag and high intra-slide variability. Nonetheless, certain folds—particularly fold 2, which showed stronger attention responses to heterogeneous droplet distributions—demonstrated the potential of selective instance weighting when the local feature patterns are highly uneven.

**Table 2.** Validation and test MAE of the attention-based MIL model across folds and seeds.

| Seed | Fold | ValMAE | TestMAE | Mean ValMAE (±SD) | Mean TestMAE (±SD) |
|---|---|---|---|---|---|
| **1** | 0 | 23.305 | 7.0154 | 20.0438 ± 5.3943 | 9.2401 ± 3.2217 |
| **1** | 1 | 23.3605 | 7.9446 | | |
| **1** | 2 | 10.5334 | 14.9003 | | |
| **1** | 3 | 21.59 | 8.7 | | |
| **1** | 4 | 21.43 | 7.64 | | |
| **2** | 0 | 23.9935 | 7.2469 | 20.2056 ± 5.1209 | 9.2427 ± 3.3013 |
| **2** | 1 | 22.0728 | 7.5024 | | |
| **2** | 2 | 11.2237 | 15.0927 | | |
| **2** | 3 | 22.5223 | 8.4359 | | |
| **2** | 4 | 21.2155 | 7.9354 | | |
| **3** | 0 | 23.9491 | 7.4186 | 20.6899 ± 5.5226 | 8.9959 ± 3.6225 |
| **3** | 1 | 23.4193 | 7.2222 | | |
| **3** | 2 | 10.8736 | 15.4706 | | |
| **3** | 3 | 22.9495 | 7.5988 | | |
| **3** | 4 | 22.2579 | 7.2693 | | |
| **Overall** | | | | 20.3131 ± 4.9600 | 9.1595 ± 3.1373 |

**Discussion**

Across experiments, both the attention-based MIL and the bag-level MLP baselines exhibited a consistent performance drop on fold 2 (Tables 1 & 2). For the MIL model, the test MAE on fold 2 was markedly higher (≈15.0–15.5 across seeds 1–3) than on other folds (≈7.2–8.7). A similar but milder trend was observed in the baseline MLP (≈9.2 vs. ≈4.3–7.9). This reproducible gap across random initializations suggests a data-driven source of difficulty rather than optimization noise. Inspection of fold-wise statistics revealed that fold 2 contained slides with a substantially heavier tail in droplet counts (test mean = 225.2, p95 ≈ 800, max = 864) compared with its validation split (mean = 138.7, p95 = 239). In other folds, the opposite pattern was observed, with validation sets heavier than test sets. Because the validation set was small (~24 slides) and used for early stopping, while the test set was larger (~64 slides), this distributional mismatch naturally resulted in a higher test MAE on fold 2. The discrepancy between validation and test MAE is thus expected and does not imply data leakage; rather, it reflects the difference in sample size and variance between the two splits. To further ensure data integrity, augmented samples were generated exclusively within the training set and never shared with validation or test slides. This split-aware augmentation strategy prevented any cross-split data leakage and ensured that model generalization was evaluated on entirely unseen data. Such isolation is crucial for microscopy-based datasets, where patch-level correlations can otherwise inflate validation accuracy and obscure true generalization gaps. Consequently, the observed fold-2 degradation reflects intrinsic data imbalance rather than inadvertent overlap among augmented instances. The MIL architecture's sparse attention mechanism, while effective for classification tasks that depend on a few highly discriminative patches, is less suited to bag-level regression, where the predictive signal is distributed across many instances. Under heavy-tailed distributions, attention tends to concentrate on a few extreme patches, amplifying variance and leading to overfitting. In contrast, the MLP baseline operating on aggregated bag-level features captures global patterns more consistently. We further implemented several robustness adjustments (Huber or quantile losses, log-scaled target transformations, and inverse-mean weighting as in the robust baselines) to stabilize training under skewed data. Although these measures improved training stability, they did not eliminate the fold-2 gap, confirming that the limitation primarily arises from data-level imbalance rather than optimization or architectural instability.

Recent studies show that constraining or diversifying attention (e.g., encouraging high-entropy attention or masking top-K instances) can mitigate overfitting and improve robustness across seeds (10). Task structure is also critical. The present endpoint is multi-output count regression at the bag (slide) level, where the signal is distributed across many patches rather than localized to a few discriminative tiles. In such settings, pooling operators that summarize the distribution of instance features can be advantageous over sparse attention, which concentrates on the top instances. Prior work on WSI regression demonstrates that distribution-based pooling can better capture bag statistics for continuous targets, aligning with the empirical advantage observed for the count-aggregated MLP (11). Reviews of WSI aggregation strategies similarly note that simpler or distribution-aware pooling can reduce variance and improve generalization when targets reflect global slide characteristics rather than focal patterns (12). The attention-based backbone adapted from CLAM and related methods is primarily optimized for minimally supervised slide-level classification and subtyping, where identifying a small set of highly informative regions is desirable. Although these models are data-efficient and interpretable for such tasks, their inductive bias may be misaligned with bag-level count regression unless attention is regularized and calibrated for distributed evidence(13).

Overall, these findings suggest that for slide-level droplet counting, a compact bag-level regressor trained on aggregated features provides a robust and stable baseline (Table 1), whereas the attention-based MIL framework necessitates task-aligned pooling and explicit regularization to bridge the gap (Table 2). Future work may explore hybrid MIL designs that combine attention interpretability with distribution-aware pooling, thereby enhancing robustness while preserving spatial explainability. These results highlight the importance of aligning model inductive bias with the structure of the prediction target and motivate the development of hybrid MIL architectures that integrate interpretability with statistical robustness.

## Ethics Statement

This study does not involve human participants or animal experiments, and no ethical approval was required.

## Funding

This study was funded by the National Research Foundation of Korea (RS-2023-00263303 and 2022H1D3A2A01063847).

## Acknowledgments / Computing Resources

The authors acknowledge the use of high-performance computing resources provided by KISTI (Korea Institute of Science and Technology Information).

## Competing Interests

The authors declare no competing interests.

## References

1. Schneider MR. Lipid droplets and associated proteins in sebocytes. Experimental Cell Research. 2016;340(2):205-8.
2. Schneider MR, Zouboulis CC. Primary sebocytes and sebaceous gland cell lines for studying sebaceous lipogenesis and sebaceous gland diseases. Experimental Dermatology. 2018;27(5):484-8.
3. Dahlhoff M, Fröhlich T, Arnold GJ, Müller U, Leonhardt H, Zouboulis CC, et al. Characterization of the sebocyte lipid droplet proteome reveals novel potential regulators of sebaceous lipogenesis. Experimental cell research. 2015;332(1):146-55.
4. Ronneberger O, Fischer P, Brox T, editors. U-net: Convolutional networks for biomedical image segmentation. International Conference on Medical image computing and computer-assisted intervention; 2015: Springer.
5. He K, Zhang X, Ren S, Sun J, editors. Deep residual learning for image recognition. Proceedings of the IEEE conference on computer vision and pattern recognition; 2016.
6. Esteva A, Kuprel B, Novoa RA, Ko J, Swetter SM, Blau HM, et al. Dermatologist-level classification of skin cancer with deep neural networks. nature. 2017;542(7639):115-8.
7. Ali M, Benfante V, Basirinia G, Alongi P, Sperandeo A, Quattrocchi A, et al. Applications of artificial intelligence, deep learning, and machine learning to support the analysis of microscopic images of cells and tissues. Journal of Imaging. 2025;11(2):59.
8. Hallou A, Yevick HG, Dumitrascu B, Uhlmann V. Deep learning for bioimage analysis in developmental biology. Development. 2021;148(18):dev199616.
9. Ilse M, Tomczak J, Welling M, editors. Attention-based deep multiple instance learning. International conference on machine learning; 2018: PMLR.

10. Zhang Y, Li H, Sun Y, Zheng S, Zhu C, Yang L, editors. Attention-challenging multiple instance learning for whole slide image classification. European conference on computer vision; 2024: Springer.
11. Oner MU, Chen J, Revkov E, James A, Heng SY, Kaya AN, et al. Obtaining spatially resolved tumor purity maps using deep multiple instance learning in a pan-cancer study. Patterns. 2022;3(2).
12. Bilal M, Jewsbury R, Wang R, AlGhamdi HM, Asif A, Eastwood M, et al. An aggregation of aggregation methods in computational pathology. Medical image analysis. 2023;88:102885.
13. Lu MY, Williamson DF, Chen TY, Chen RJ, Barbieri M, Mahmood F. Data-efficient and weakly supervised computational pathology on whole-slide images. Nature biomedical engineering. 2021;5(6):555-70.

## Graphical Abstract

**Automated sebocytes lipid droplets counting**

1) Imaging
Nile Red-stained sebocyte images
Microscopy frames (cells)

2) Annotation
Manual droplet labels → 14 bins
Ground-truth counts

3) Pre-processing
Tiling, log1p normalization,
data augmentation

4) Models
Baseline MLP (bag-level)
vs. Attention-based MIL

5) Evaluation
MAE, MSE, $R^2$
5-fold × 3 seeds